\documentclass{ifacconf}

\makeatletter
\let\old@ssect\@ssect
\makeatother

\usepackage{natbib}
\usepackage{graphicx}      
\usepackage{amsmath}
\usepackage{algorithm}
\usepackage{algpseudocode}
\usepackage{tikz}
\usepackage{tikz-qtree}
\usepackage{float}
\usepackage{makecell}
\usepackage{comment}
\usetikzlibrary{positioning,fit,calc,arrows,shapes}
\algnewcommand{\LineComment}[1]{\State \(\triangleright\) #1}
\usepackage[colorlinks=true, allcolors=blue, bookmarks=false]{hyperref}

\makeatletter
\def\@ssect#1#2#3#4#5#6{%
  \NR@gettitle{#6}
  \old@ssect{#1}{#2}{#3}{#4}{#5}{#6}%
}
\makeatother

\begin{document}
\begin{frontmatter}

\title{Discovery of fully efficient fault indicators along a data-based diagnosis process} 

\author[First]{Igor Bezmaternykh} 
\author[First]{Louise Travé-Massuyès} 
\author[First]{Elodie Chanthery}

\address[First]{LAAS-CNRS, Université de Toulouse, INSA, Toulouse, France e-mails: \tt bezmaternykh@insa-toulouse.fr, louise@laas.fr, echanthery@laas.fr}

\begin{abstract}                
The integration of model-based and data-driven paradigms provides a powerful framework for fault diagnosis by combining the interpretability of analytical redundancy relations, i.e., input-output relations that are used as diagnosis indicators in model-based diagnosis, with the adaptability of learning techniques. DT4X is a recent diagnosis algorithm that  uses symbolic regression to generate multivariate relations leveraging some properties of analytical redundancy relations and uses them as split functions in a decision tree. However, its symbolic regression procedure optimizes only the separation between two selected classes at each node, often fragmenting the remaining classes and degrading both interpretability and diagnosis performance. This paper introduces DT4X+, an enhanced version of DT4X that modifies the construction of training sets and the symbolic-regression loss so that expressions separate the target classes while preserving the coherence of non-target classes. The resulting relations become fully consistent with ARR properties and lead to more informative splits, improved robustness, and better performance on dynamic-system datasets. Experiments conducted on several benchmark systems demonstrate the benefits of this enhanced formulation.
\end{abstract}

\begin{keyword}
Data-based diagnosis, knowledge discovery, diagnosis indicators, decision trees
\end{keyword}

\end{frontmatter}

\section{Introduction}
Data-based diagnosis methods are increasingly popular, as they enable fault diagnosis even when expert knowledge is limited or when a method must be rapidly adapted to a new system without redesigning a full model. However, black-box learning approaches, such as neural networks, often lack interpretability, which is a key requirement in fault diagnosis. This motivates the use of decision trees, whose structure and split functions bring insights into the diagnostic process. Several studies ~(\cite{assaf2005build, gaddam2007k,guh2008effective,sun2007decision}) have applied decision trees to fault detection and isolation. Yet, traditional univariate trees remain inadequate because, in model-based diagnosis, the distinction between normal and faulty behaviours relies on multivariate relations, which can be captured by analytical redundancy relations (ARRs) when an analytical model is known, rather than by single-variable relations.

This limitation has motivated the use of multivariate decision trees (\cite{canete2021review}). Some contributions rely on linear combinations of the variables used as features, while others incorporate non-linear split functions using synthetic features, kernel methods, or neural networks as presented in Section~\ref{related_work} that surveys related work. Although these approaches increase expressiveness, they often require strong prior knowledge about the expected functional form. More recently, symbolic-regression-based decision trees have emerged as a promising alternative, as they can discover meaningful multivariate relations automatically~(\cite{fong2024symbolic}). Among them, DT4X (\cite{goupil2024tree}) has shown that symbolic regression can be guided to produce expressions leveraging some properties of analytical redundancy relations ARRs, thus reconnecting data-driven learning with well-established concepts from model-based diagnosis.

Despite its strengths, DT4X has an intrinsic limitation. Its symbolic-regression procedure optimizes the discrimination between two target classes at each node but does not impose the behavior of the resulting expression for the remaining classes. As a result, non-target classes may be arbitrarily fragmented across child nodes, producing split functions that are less informative and forcing the tree to compensate with additional relations at deeper levels. In this respect, the identified relationships do not fully align with the properties of ARRs, since an ARR partitions all classes entirely into two distinct groups. As a matter of fact, ARRs are zero for samples of the nominal class. For other classes, ARRs may be zero (when not sensitive to the fault of the class) or non-zero (when sensitive to the fault of the class), and they remain consistent across all samples within a given class. Consequently, classes can be discriminated using a sufficient set of ARRs and the corresponding sensitivity Boolean vector, known as the ARR fault support.

To address this issue, we propose DT4X+, an enhanced version of DT4X that incorporates all classes during symbolic regression and introduces a loss term penalizing the fragmentation of non-target classes. This modification encourages the discovery of split functions that better preserve class structure while still separating the target classes. The resulting relations are then fully consistent with ARR properties, serving as fully efficient fault indicators, and ultimately yield more coherent and interpretable decision trees.

The article is organized as follows: Section 2 reviews related work on multivariate and symbolic-regression-based decision trees. Section 3 presents the DT4X algorithm. Section 4 introduces the DT4X+ extension and its revised loss function. Section 5 reports experimental results on both static and dynamic systems. Finally, section 6 concludes the paper and outlines perspectives for future work.

\section{Related work}\label{related_work}

Multivariate decision trees have been explored through a variety of formulations over the years. Early approaches focused on linear combinations of features to define oblique splits, as in OC1 and its variants (\cite{murthy1993oc1,CO2algorithm}), or more recent continuous-optimization methods (\cite{utgoff1989perceptron,wickramarachchi2016hhcart}).
While these approaches improve expressiveness compared to univariate trees, their linear nature limits their ability to capture the nonlinear relations typically required for fault indicators.

Several strategies have been proposed to address this limitation. Some works augment the feature space by generating synthetic attributes through algebraic combinations of original variables (\cite{kou2020data}), enabling standard tree learners to exploit nonlinear relationships. However, such feature expansion scales poorly when the number of variables increases, often leading to combinatorial explosion. Other methods (\cite{hutchison2024active}) incorporate a limited set of predefined nonlinear split functions within otherwise univariate trees. These techniques rely heavily on prior domain knowledge and restrict the variety of relations that can be discovered.

A different class of methods embeds nonlinear classifiers within the nodes of a decision tree. Approaches based on support vector machines (\cite{bennett1998support} and \cite{montanana2021stree}) make use of kernel functions to obtain nonlinear boundaries, whereas neural or perceptron trees (\cite{balestriero2017neural, yildiz2001omnivariate}) rely on multilayer perceptrons to model split functions. While these produce flexible decision boundaries, the resulting models lose interpretability due to the opaque nature of their internal parametrization, which is undesirable for fault diagnosis.

Symbolic-regression-based trees offer an alternative that preserves interpretability while enabling nonlinear multivariate splits. The SREDT algorithm proposed by \cite{fong2024symbolic} uses genetic programming to generate candidate expressions and selects them using a split-impurity criterion. This method performs competitively on fault-oriented datasets, but the expressions it generates do not correspond to analytical redundancy relations (ARRs), as the splitting threshold remains arbitrary.

DT4X, proposed by \cite{goupil2024tree}, addresses this limitation by constraining symbolic regression (\cite{kronberger2024symbolic}), embedded into symbolic classification, to produce expressions leveraging some properties of ARRs, in particular the fact that ARRS are (ideally) zero for the nominal class and non-zero for some faulty classes. DT4X learns such analytical relations and integrates them directly into a decision-tree structure. This approach yields accurate diagnosis trees with meaningful, interpretable split functions derived automatically from data.


However, DT4X does not fully account for all the properties of ARRs. Indeed, DT4X constructs each split by focusing solely on the discrimination between a pair of target classes. As a consequence, samples from other classes may be arbitrarily partitioned across the tree’s branches. This fragmentation can hinder interpretability and decrease diagnostic performance, especially for systems with many fault modes or heterogeneous behaviours. This is not surprising since it contravenes the definition of ARRs, whereby all samples in a class must behave in the same way (evaluating ARRs 0 or non-zero). 

DT4X+ builds upon DT4X and addresses this limitation. By incorporating all classes during symbolic regression and penalizing the fragmentation of non-target classes, DT4X+ leads to  coherent splits while fully preserving the ARR properties of the learned relations.

\section{The tree based diagnosis method DT4X}
\label{sec:DT4X_algo}

DT4X, detailed in (\cite{goupil2024tree}), is briefly reminded here. It constructs a diagnosis tree $T$ recursively, generating at each node a multivariate  relation that separates two target classes ($(c_1, c_2)$ in Algorithm~\ref{alg:DT4X}). These relationships are derived through symbolic classification, constrained to emulate a key property of ARRs: an ARR that evaluates to zero for samples of one class and non-zero for samples of another class can discriminate between the two classes. Symbolic classification subsumes symbolic regression to automatically discover explicit mathematical expressions to distinguish between classes, based on input features, i.e., variables and operators.

At each node, DT4X identifies the subset of classes that are sufficiently represented in the local dataset. If the nominal class $c_0$ is present in sufficient proportion, the algorithm considers target pairs of the form $(c_0, c_x)$ with $c_x$ one of the other sufficiently represented classes. Otherwise, all ordered pairs of sufficiently represented classes are considered (\Call{generate pairs}{}, line 9, Algorithm~\ref{alg:DT4X}). 

The learning process for a given pair proceeds as follows. First, a training set is built by sampling data from the two target classes. When a non-nominal class is involved, nominal samples are added to the training set and assigned to the first class of the pair to enforce the ARR structure. In any case, the \Call{balance}{} function creates two perfectly balanced sets of samples  (line 12, Algorithm~\ref{alg:DT4X}, and see \Call{balance}{} function in Algorithm~\ref{alg:balance}). Second, DT4X attempts to construct an analytical relation that separates the two classes. For this purpose, symbolic classification (SC, line 13, Algorithm~\ref{alg:DT4X}) is applied using a threshold around zero as the decision function and the \textproc{log\_loss} function as the fitness criterion:
\begin{equation}\label{log_loss}
log\_loss = -\frac{1}{N} \sum_{i=1}^{N} \left[ y_i \log(p_i) + (1 - y_i)\log(1 - p_i) \right]
\end{equation}
where:
\begin{itemize}
    \item $N$ is the number of samples.
    \item $y_i$ is the label (0 ou 1) for sample $i$.
    \item $p_i$ is the predicted probability for the sample $i$ to belong to class~1.
\end{itemize}

This encourages the algorithm to identify expressions that evaluate close to zero for one target class and have a non-zero value for the other. If the best expression sufficiently separates the target classes, the node is split according to the expression (lines 17 and 18, Algorithm~\ref{alg:DT4X}), and the procedure is applied recursively to the resulting child nodes. If no candidate expression achieves the required separation, the node becomes a leaf labeled with the majority class (lines 20 and 21, Algorithm~\ref{alg:DT4X}).

The tree is grown in a breadth-first manner: for each node, class pairs are evaluated sequentially until a suitable expression is found or all pairs are exhausted. The resulting decision tree contains at each internal node a symbolic expression playing the role of a fault indicator, while the leaves correspond to fault classes.

\begin{algorithm}
\caption{DT4X pseudo-code}
\label{alg:DT4X}
\textbf{Inputs:} $\mathcal{D}$, $O$, Hyper-Parameters, untrained DT ($n_0$) \\
\textbf{Output:} $T$ trained decision tree with diagnosis indicators
\begin{algorithmic}[1]
\State Initialize tree $T$ with a single root node $n_0$ containing dataset $\mathcal{D}$ 
\State Initialize $currentNodes \gets n_0$
\While{$currentNodes$ is not empty}
    \ForAll{$node \in currentNodes$}
        \If{$node$ is pure with label $l$}
            \State $node$ is leaf
            \State $node \gets l$
        \Else
            \State $pairsToTry \gets$ \textproc{generate pairs}
            \State $(c1, c2) \gets$ first element of $pairsToTry$
            \While{not $c_{best}$ and $pairsToTry$ not empty}
                \State $X_{train}, y_{train} \gets$ \Call{balance}{$\mathcal{D}_n$, c1, c2} 
                \State {\raggedright $c_{best} \gets$ \Call{\mbox{SC}}{$X_{train}, y_{train}$}}
                
                \State $pair \gets$ next element of $pairsToTry$
            \EndWhile
            \If{$c_{best}$}
                \State $lNode, rNode \gets$ split according to $c_{best}$
                \State $futureNodes \gets +lNode, rNode$
            \Else
                \State $node$ is leaf
                \State $node \gets$ majority label $l$
            \EndIf
        \EndIf
    \EndFor
    \State $currentNodes \gets futureNodes$
\EndWhile
\end{algorithmic}
\end{algorithm}


\section{Enhancing DT4X with DT4X+}

This section introduces DT4X+, an extension of DT4X designed to address the fragmentation of non-target classes during split construction. The modifications concern both the construction of the symbolic-regression training sets and the definition of the loss function used to evaluate candidate expressions.

\subsection{Identification of the problem}
\begin{figure}[h]
\label{fig:class_separation}
\centering
\begin{tikzpicture}[
  node distance=2mm,
  title/.style={font=\fontsize{10}{10}\selectfont\color{black!50}\ttfamily},
  typetag/.style={ellipse, draw=black!50, font=\scriptsize\ttfamily}
]

  \node (N0) at (1.5cm,1) [title] { Node 0 };

  \node (cl0) [below=of N0, typetag] { Nominal };
  \node (cl1) [below=of cl0, typetag] { Fault 1 };
  \node (cl2) [below=of cl1, typetag] { Fault 2 };
  \node (expr) [below=of cl2, typetag, style={rectangle, draw, minimum width=2.8cm,minimum height=4mm, inner sep=0pt}] { Symbolic expression };
  \coordinate (N0out) at ($(expr.south) + (0,0)$);

  \node [draw=black!50, fit={(N0) (cl0) (cl1) (cl2)}] {};

  \node (N1) at (0, -3cm) [title] { Node 1 };

  \node (cl10) [below=of N1, typetag] { Nominal };
  \node (cl11) [below=of cl10, typetag] { \textcolor{red}{Fault 1} };
  \node [draw=black!50, fit={(N1) (cl10) (cl11)}] {};

  \node (N2) at (3cm, -3cm) [title] { Node 2 };

  \node (cl21) [below=of N2, typetag] {Fault 2};
  \node (cl22) [below=of cl21, typetag] { \textcolor{red}{Fault 1} };

  \node [draw=black!50, fit={(N2) (cl21) (cl22)}] {};
  
  \draw[->] (expr.south) -- (N1);
  \draw[->] (expr.south) -- (N2);
\end{tikzpicture}
\caption{Illustration of a non-target class being fragmented.}
\end{figure}
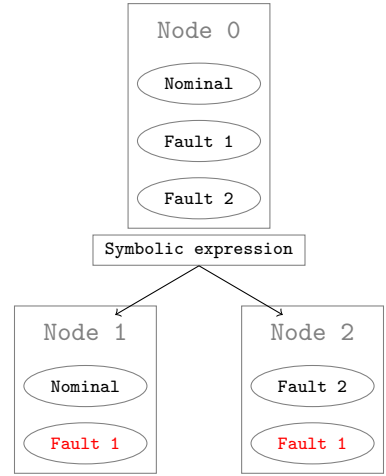

In DT4X, the symbolic-regression process optimizes expressions exclusively to separate a selected pair of target classes. Samples belonging to other classes do not contribute to the loss and therefore do not influence the structure of the learned relation. As a consequence, an expression that cleanly separates the two target classes may arbitrarily divide the remaining classes across the two branches of the split. This phenomenon is illustrated in Fig.~\ref{fig:class_separation}: although the split correctly separates the target pair (Nominal and Fault 2), it fragments another class (Fault 1) across both children, compromising interpretability and forcing the tree to compensate with additional relations at deeper levels.

This behaviour is problematic for two reasons. First, a relation that fragments a class conveys no diagnosis information about that class, contradicting the ARR-inspired objective of producing interpretable indicators. Second, unnecessary fragmentation increases the number of required splits, leading to deeper and potentially less robust trees. These limitations become more pronounced in datasets with many fault classes or heterogeneous fault signatures. 

\subsection{DT4X+ revised loss}
\label{sec:rev_loss}
To mitigate class fragmentation, DT4X+ introduces an additional loss term that penalizes splits responsible for dispersing non-target classes across different branches. The symbolic classification objective therefore becomes a combination of two components:
(1) a discrimination loss, denoted $L_{\textrm{target}}$ that enforces discrimination of the target pair, identical to the DT4X loss given by the $log\_loss$ of equation \eqref{log_loss}, and
(2) a fragmentation loss, denoted $L_{\textrm{frag}}$, that measures how non-target classes are distributed after the split:
\begin{equation}
Loss_{DT4X^+}= L_{\textrm{target}}+ L_{\textrm{frag}} 
\end{equation}

For the fragmentation term, DT4X+ uses the gini impurity computed separately for each non-target class.  This split score is similar to the metric of split impurity commonly used in decision trees, but calculated over classes instead of over nodes. It is more coherent because it aligns the impurity measure directly with the distribution of each class within a node, rather than aggregating across all classes indiscriminately. Let $nt$ be the set of non-target classes, the fragmentation loss, using the gini impurity normalized for a binary split, is defined as:
\begin{equation}
   L_{\textrm{frag}}= \sum\limits_{k \in nt}4\cdot p_{\textrm{left},k}\cdot p_{\textrm{right},k}\cdot w_k
\end{equation}
with $p_{\textrm{left,k}}$  and $p_{\textrm{right},k}=1-p_{\textrm{left},k}$ the proportions of samples of class $k$ sent to each branch, and $w_k$ the weight of class $k$ according to sample count, i.e., its relative frequency (lines 11-15, Algorithm~\ref{alg:loss}).  $L_{\textrm{frag}}$ is essentially a weighted and normalized version of the Gini impurity applied to each non-target class. The loss for one class $k$ is therefore 0 when all samples of class $k$ fall into a single branch (no fragmentation) and 1 when the class is split evenly between both branches (highest fragmentation). The factor "4" normalizes the Gini impurity for a binary split so that its maximum value is 1 (when $p_{\text{left},k} = p_{\text{right},k} = 0.5$). By summing over all non-target classes, we obtain a global measure of undesirable fragmentation caused by the split. This loss is small when each non-target class remains entirely in a single branch (good separation) and large when non-target classes are dispersed between the two branches (high fragmentation). Hence, it favors splits that preserve the integrity of non-target classes.

The metric $L_{\textrm{frag}}$ is simple to compute as it does not require exponentiation or logarithms. This is important because it is calculated on all samples of non-target classes. When combined with the log-loss used for target-class discrimination, it ensures that fragmentation is only penalized when the expression already achieves sufficiently high separation accuracy on the target pair, thus preserving the target pair discrimination behaviour, as detailed below.


The \textproc{log\_loss} function (line 5, Algorithm~\ref{alg:loss}) cannot operate on probabilities equal to~0, since 
$\log(0)$ is undefined. To avoid this issue, the \texttt{gplearn} implementation that we use adopts \emph{clipping}, replacing 
predicted probabilities equal to~0 with $10^{-15}$, and probabilities equal to~1 with 
$1 - 10^{-15}$. In DT4X, the decision function outputs only $0$ or $1$, so after clipping, the 
effective probabilities passed to \textproc{log\_loss} are restricted to the set 
$\{10^{-15},\, 1 - 10^{-15}\}$. As a consequence, the log-loss of a sample takes only two 
possible values, depending solely on whether the sample is correctly or incorrectly classified.

For a misclassified sample, the clipped probability is $p = 10^{-15}$, leading to a log-loss of
\[
    -\log(10^{-15}) = 15 \log(10) \approx 34.54.
\]
For a correctly classified sample, the clipped probability is $p = 1 - 10^{-15}$, for which
\[
    -\log(1 - 10^{-15}) \approx 10^{-15},
\]
a negligible value. Thus, the total log-loss becomes essentially proportional to the number of 
misclassified samples. Being $N$ the total number of samples and $N_{\text{incorrect}}$ the number of 
misclassified ones, the accuracy is defined as
\[
    \text{accuracy} = 1 - \frac{N_{\text{incorrect}}}{N}.
\]
Since each misclassified sample contributes approximately 15 $\log(10)$ to the log-loss, we obtain
\[
    \text{log\_loss} \approx (1 - \text{accuracy}) \times 15 \log(10).
\]
The $L_{\textrm{frag}}$ value lies within [0,1], so it has an influence on $L_{DT4X^+}$, the total loss of DT4X+, only when $L_{target}$ is less than 1.  
\begin{align*}
L_{target}=\text{log\_loss} < 1 
&\;\;\Leftrightarrow\;\; (1 - \text{accuracy}) \times 15 \log(10) < 1 \\
&\;\;\Leftrightarrow\;\; \text{accuracy} > 0.97.
\end{align*}

The threshold of 0.97 for accuracy corresponds to a high target-class separation quality, confirming that $L_{\textrm{frag}}$ enters into play only when sufficient high separation of the target classes is achieved.

\subsection{DT4X+ algorithm}

DT4X+ modifies two parts of the original algorithm: the construction of the training dataset and the computation of the fitness function. Unlike DT4X, the training dataset always includes all classes present in the node. Target classes are balanced as before, while all other classes are added in their original proportions. This ensures that symbolic regression receives information about how non-target classes respond to candidate expressions. The modified \textproc{balance} function is described in Algorithm~\ref{alg:balance}.

During training, the fitness of a candidate expression becomes the sum of the DT4X classification loss on the target classes $L_{\textrm{target}}$ and the fragmentation loss on the non-target classes $L_{\textrm{frag}}$ (Algorithm~\ref{alg:loss}). Expressions that cleanly separate the target pair while maintaining class coherence are therefore favoured.

\begin{algorithm}
\caption{DT4X vs DT4X+ \textproc{balance} pseudo code - The text in magenta shows the changes made in DT4X+ compared to DT4X in black.}
\label{alg:balance}
\textbf{Inputs:} $\mathcal{D}_n$: dataset at current node, $(c1,c2)$: target pair classes\\
\textbf{Output:} $X_{train}, y_{train}$: training dataset for SC (symbolic classifier)
\begin{algorithmic}[1]
\State $N_{min} \gets$ min(count($\mathcal{D}_n$, c1), count($\mathcal{D}_n$, c2))
\LineComment{Build samples for class c2}
\State $S_{c2}\gets$ \Call{sample}{$\mathcal{D}_n$, c2, $N_{min}$}
\If{$c1 \neq nominal$}
    \State $S_{nom} \gets$ \Call{sample}{$\mathcal{D}_n$, nominal, $N_{min}/2$}
    \State $S_{c1} \gets$ \Call{sample}{$\mathcal{D}_n$, c1, $N_{min}/2$}
    \State $S_{c1} \gets$ \Call{concatenate}{$S_{nom},S_{c1}$} 
\Else
    \State $S_{c1} \gets$ \Call{sample}{$\mathcal{D}_n$, c1, $N_{min}$}
\EndIf
\LineComment{\textcolor{magenta}{Add non-target classes in original proportions}}
\State \textcolor{magenta}{$S_{extra} \gets$ samples in $\mathcal{D}_n$ whose labels are not in $\{nominal, c1, c2\}$}
\State $X_{train} \gets$ \Call{concatenate}{$S_{c1}$, $S_{c2}$, \textcolor{magenta}{$S_{extra}$}}
\State $y_{train} \gets$ \Call{labels}{$X_{train}$}
\State \Return \Call{shuffle}{$X_{train}, y_{train}$}
\end{algorithmic}
\end{algorithm}


\begin{algorithm}
\caption{DT4X+ \textproc{loss} pseudo code}
\label{alg:loss}
\textbf{Inputs:} $\hat{y}$: predicted class labels from SC, $y$: true class labels, (c1, c2): target class pair
\begin{algorithmic}[1]
\LineComment{Target-class loss (same as DT4X)}
\State mask\_target $\gets$ indices where $y \in \{c1, c2\}$
\State $\hat{y}_{target} \gets \hat{y}$[mask\_target]
\State $y_{target} \gets y$[mask\_target]
\State $L_{target} \gets$ \Call{log\_loss}{$\hat{y}_{target},y_{target}$}
\LineComment{Fragmentation loss on non-target classes}
\State mask\_non\_target $\gets$ indices where $y \notin \{c1, c2\}$
\State $\hat{y}_{nt} \gets \hat{y}$[mask\_non\_target]
\State $y_{nt} \gets y$[mask\_non\_target]
\State $L_{frag}  \gets 0$
\ForAll{class $k$ in \Call{unique}{y\_{nt}}}
    \State $p_{left} \gets$ proportion of samples of class $k$ with $\hat{y}_{nt} = left$
    \State $p_{right} \gets$ proportion of samples of class $k$ with $\hat{y}_{nt} = right$
    \State $w_k$ $\gets$ \Call{count}{$y_{nt}==k$}/\Call{length}{$y_{nt}$}
    \State $L_{frag} \gets L_{frag} + 4 \cdot p_{left} \cdot p_{right} \cdot w_k$
\EndFor
\State \Return $L_{target} + L_{frag}$
\end{algorithmic}
\end{algorithm}

For all other aspects such as pair selection, recursive tree construction, stopping criterion, DT4X+ follows the same procedure as DT4X (\cite{goupil2024tree}). 

\section{Experiments}
\subsection{Experimental Setup}
\textit{Datasets --}
DT4X+ was evaluated on four datasets used for fault diagnosis: a static polybox system composed of adders and multipliers, a static logical circuit representing a subtractor (\cite{goupil2023tree}), a dynamic two-tank system with 12 different possible faults (\cite{goupil2024tree}) and a cyber-attack detection dataset (\cite{syfert2023control}) corresponding to the superheater 4.2 subsystem of a steam-generation process. 
For the dynamic datasets, additional temporal features were generated using finite-difference approximations (NumPy’s gradient operator) to capture dynamic behaviour. For the superheater dataset, the training set was subsampled to reduce training time.

\textit{Baselines --}
DT4X was taken as the main baseline for evaluation, as it provides state-of-the-art results on these problems and shares the same symbolic-regression backbone as DT4X+ (\cite{goupil2023tree}). Using DT4X as a reference enables a direct comparison of the diagnostic relevance and interpretability of the relations produced by both methods.

\textit{Evaluation Metrics --}
Performance was assessed using accuracy and F1-score, which are suitable for class-imbalanced fault datasets. Computational cost was evaluated through training time and inference latency. This allows analysing the benefit–cost trade-off introduced by DT4X+.

\textit{Implementation Details --}
Both algorithms, DT4X and DT4X+, were implemented in Python using the $gplearn$ library for genetic-programming-based symbolic regression. Some $gplearn$ classes were overridden to reduce overhead and ensure consistent behaviour across both methods. Experiments were run on an HP 996M4ET workstation. To achieve a fair comparison, the two algorithms share the same hyperparameter values.

The code is made available on GitHub\footnote{\href{https://github.com/Igor-Bzk/DT4X_plus}{https://github.com/Igor-Bzk/DT4X\_plus}} as well as the hyperparameter values.

\subsection{Evaluation Results}

Tables 1 and 2 report the performance obtained by DT4X and DT4X+ on all datasets.

\label{sec:results}
\begin{table}[H]
    \resizebox{\columnwidth}{!}{%
    \begin{tabular}{|c|c|c|c|c|l|}\hline
         \textbf{Dataset}&  \textbf{Accuracy}&  \textbf{F1-scores}&  \textbf{Inference Time}& \textbf{Training time}\\\hline
         sf\_polybox& 0.79909& 0.73210& 0.00067s & 158.15s\\\hline
         sf\_subtractor& 0.84989& 0.82223& 0.0019s& 240.32s\\\hline
         sw\_water\_tanks& 0.99713 & 0.99713& 0.022s& 1909.62s\\\hline
         steam\_superheater\_42& 0.98968 & 0.98983 & 0.0013s & 96.67s\\\hline
    \end{tabular}
    }
    \caption{DT4X results per dataset}
    \label{tab:DT4X_results}
\end{table}

\begin{table}[H]
    \resizebox{\columnwidth}{!}{%
    \begin{tabular}{|c|c|c|c|c|l|}\hline
         \textbf{Dataset}&  \textbf{Accuracy}&  \textbf{F1-scores}&  \textbf{Inference Time}& \textbf{Training time}\\\hline
         sf\_polybox& 0.79909& 0.73210& 0.00075s & 188.26s\\\hline
         sf\_subtractor& 0.84867 & 0.81647 & 0.0018s & 546.19s\\\hline
         sw\_water\_tanks& 0.99847 & 0.99847 & 0.031s & 10318.46s\\\hline
         steam\_superheater\_42& 0.99479 & 0.99483 & 0.0015s & 187.91s\\\hline
    \end{tabular}
    }
    \caption{DT4X+ results per dataset}
    \label{tab:DT4X+_results}
\end{table}

On the polybox system, both algorithms produced identical trees, resulting in identical accuracy and F1-scores. On the subtractor circuit, DT4X+ shows a slight decrease in global metrics (–0.012 in accuracy, –0.058 in F1-score). However, a detailed inspection reveals that DT4X fails to distinguish classes 1 and 2, whereas DT4X+ successfully isolates a subset of class 1 without introducing false positives. The improved behaviour is confirmed by the confusion matrices and tree structures (cf. Github\footnotemark[1]). 
DT4X+ provides a consistent improvement on dynamic datasets. On the two-tank system, both accuracy and F1-score increase by around 0.02. On the superheater dataset, the gains reach approximately 0.05. These improvements reflect the ability of DT4X+ to produce relations that preserve the coherence of non-target classes, thereby enabling more meaningful splits and reducing error propagation deeper in the tree.
Inference latency remains essentially unchanged. Increases observed on the two-tank system (0.022 s to 0.031 s) are attributed to differences in tree structure rather than computational complexity. DT4X+ produced a root split that grouped more classes on one side, requiring additional downstream splits.
On dynamic datasets, DT4X+ consistently produces expressions that better separate classes and correspond more closely to ARR-like behaviours. For example, in the two-tank system, the root relation produced by DT4X fragments several classes, whereas DT4X+ isolates each class into one of the two child nodes, with only a few classes appearing in both nodes, and even then, they exhibit a negligible number of samples in one of them (cf. Github\footnotemark[1]).
The learned relations therefore provide more usable diagnostic insight.
DT4X+ isolated each of the classes into one of the two nodes, with only classes 0, 3, 5 and 10 having a presence in both Node 1 and 2, though still having a very significant majority in one of the two nodes.

\subsection{Impact on computation time}
DT4X+ and DT4X have almost identical inference time. However, training time increases significantly with DT4X+. Because the fragmentation loss requires evaluating the distribution of all non-target classes, the computational cost grows with both the number of classes and the number of samples per class. On the two-tank system, training time is multiplied by nearly 5. On the subtractor and superheater datasets, training time roughly doubles. On the polybox system, the impact remains below 20\%, due to the small number of classes and limited dataset size.

If training time matters, DT4X+ is particularly advantageous for dynamic systems, where the performance gains justify the additional training cost. For simpler static systems, the original DT4X remains a competitive option.

\section{Conclusion}

This paper proposes DT4X+, an enhanced version of the symbolic-regression-based decision tree algorithm DT4X.
The approach addresses a structural limitation of DT4X: the fragmentation of non-target classes caused by optimizing splits solely for target-pair discrimination. By modifying both the training-set construction and the symbolic-regression loss function, DT4X+ explicitly incorporates non-target classes and penalizes their dispersion across branches. The resulting split functions remain consistent with ARR principles while providing more coherent and informative class separation. They gain in interpretability because their sensitivity is defined for every class, whereas it was uncertain for some classes in the fault indicators of DT4X. Experimental results on both static and dynamic systems show that DT4X+ improves diagnostic performance, especially on dynamic datasets, where class interactions and temporal behaviours are more complex. These benefits come at the cost of a higher training time, whose magnitude depends on the number of classes and samples.
Future work will focus on integrating model-based constraints more deeply into the symbolic-regression process. This could strengthen the link between data-driven learning and diagnosis meta-knowledge.

\begin{ack}
The authors would like to thank Louis Goupil, co-designer and developer of DT4X and now Technical Lead at ATOS, for his invaluable help on DT4X. \\
This work has benefited from the AI Interdisciplinary Institute ANITI funded by the France 2030 program under the Grant agreements n°ANR-19-PI3A-0004 and n°ANR-23-IACL-0002.
\end{ack}

\bibliography{ifacconf}             


\end{document}